\documentclass[preprint]{ceurart}

\usepackage{graphicx}
\usepackage{amsmath}
\usepackage{booktabs}
\usepackage{url}
\usepackage{multirow}
\conference{Preprint}

\title{Retrieval-Guided Generation for Safer Histopathology Image Captioning}

\author[1,2]{Md. Enamul Hoq}
\author[1]{Wataru Uegami}
\author[1]{Saghir Alfasly}
\author[1]{Ghazal Alabtah}
\author[3]{Sahar Rahimi Malakshan}
\author[4]{Armita Kazemi}
\author[5]{Alex T. Schmitgen}
\author[2]{Fred Prior}
\author[1]{H.R. Tizhoosh}

\cormark[1]
\fnmark[1]
\address[1]{Kimia Lab, Department of Artificial Intelligence \& Informatics, Mayo Clinic, Rochester, MN, USA}
\fnmark[1]
\address[2]{Department of Biomedical Informatics, University of Arkansas for Medical Sciences, Little Rock, AR, USA}
\fnmark[1]
\address[3]{Lane Department of Computer Science and Electrical Engineering, West Virginia University, Morgantown, WV, USA}
\fnmark[1]
\address[4]{Department of Computer Science and Engineering, Princeton University, Princeton, NJ, USA}
\fnmark[1]
\address[5]{Department of Computer Sciences, University of Wisconsin--Madison, Madison, WI, USA}

\begin{document}

\begin{abstract}
Generative vision–language models can produce fluent medical image captions but remain prone to hallucination, over-specific diagnostic claims, and factual inconsistency—serious issues in pathology. We investigate retrieval-guided generation (RGG) as a safer alternative, where captions are formed by summarizing expert text from visually similar cases rather than generated de novo. On the ARCH histopathology dataset, RGG improves semantic alignment with ground truth, achieving cosine similarity of $\approx$0.60 versus $\approx$0.47 from MedGemma, with non-overlapping confidence intervals indicating a robust gain. A pathologist-led qualitative review shows better preservation of morphology-relevant terminology and fewer unsupported diagnoses, while revealing failure modes such as concept mixing and inherited over-specific labeling. Overall, retrieval-guided captioning offers a more transparent and reliable approach with clearer opportunities for auditing than fully generative methods.
\end{abstract}


\maketitle

\section{Introduction}

Automated captioning of histopathology images has the potential to support education, reporting, search, and clinical decision support~\cite{kalra2020yottixel, hoq2025multimodal, tizhoosh2018aichallenges}. Recent multimodal foundation models can generate descriptive captions directly from image pixels. However, direct generation in medical contexts remains problematic: fluent outputs may still be clinically incorrect. In pathology, a plausible but unsupported statement can be more harmful than a vague description, particularly when a model assigns a specific disease label without sufficient visual evidence. This challenge motivates a retrieval-guided alternative. Rather than generating a caption from scratch, we first retrieve visually similar images from an expert-annotated archive and then summarize their captions into a concise description~\cite{tizhoosh2018aichallenges}. This design grounds the output in established expert language and provides an interpretable path from image to text. It is particularly well suited to pathology, where morphology-rich descriptors recur across related cases and expert-authored captions constitute a valuable source of clinically meaningful language. Prior work in computational pathology has emphasized semantic retrieval and representation learning as key mechanisms for improving the interpretability and searchability of pathology archives~\cite{tizhoosh2021searching}. At the same time, foundation models such as MedGemma~\cite{sellergren2025medgemma} have demonstrated the promise of direct image-to-text generation. Our work bridges these directions by examining whether captioning can be made safer and more clinically grounded when generation is constrained by retrieved expert knowledge rather than performed directly from pixels. The intuition behind retrieval-guided generation is straightforward: pathology captions encode clinically meaningful morphology that has already been curated by experts. If a system can reliably retrieve visually and semantically related examples, the language model can focus on synthesizing and harmonizing existing evidence rather than generating novel claims. This shift may reduce hallucination risk while preserving fluency, effectively transferring the burden of correctness from unconstrained generation to evidence-grounded summarization. In this paper, we evaluate retrieval-guided captioning on the ARCH dataset~\cite{gamper2021arch} and compare it with a direct generative baseline, MedGemma. We further include a pathologist-guided qualitative review to assess whether quantitative improvements correspond to clinically meaningful outputs. The main contributions of this work are a retrieval-guided captioning framework for histopathology images, a quantitative comparison between direct generation and retrieval-grounded summarization using BioBERT cosine similarity \cite{lee2020biobert}, a comparison of retrieval backbones demonstrating the advantage of stronger visual embeddings, and a single-blind pathologist-guided qualitative evaluation highlighting both strengths and clinically relevant failure modes.

\section{Methodology}
\begin{figure}[ht]
\centering
\includegraphics[width=0.99\linewidth]{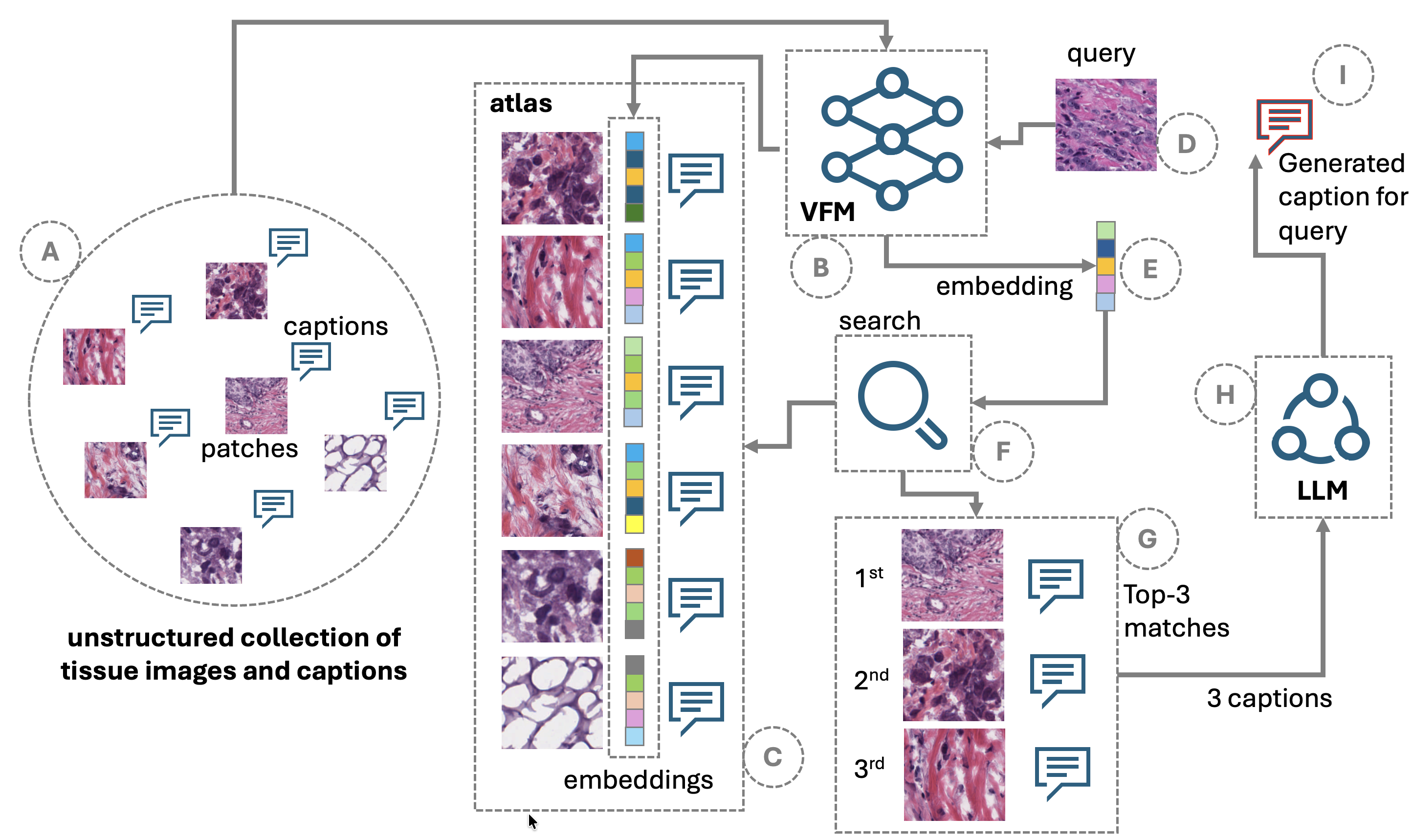}
\caption{Overview of the retrieval-guided generation (RGG) pipeline: (A–C) unstructured image–caption pairs are encoded by a vision foundation model (VFM) into embeddings that form a searchable atlas; (D–F) a query image is encoded with the same VFM and used to retrieve similar cases; (G–I) the top-3 images and their captions are passed to a large language model, which aggregates them into a final caption.}
\label{fig:pipeline}
\end{figure}

Rretrieval-guided generation (RGG)  decouples visual retrieval from language generation. Instead of generating text directly from a query image, it first retrieves visually similar images and then leverages their expert-provided captions as the foundation for aggregation and summarization (see Figure \ref{fig:pipeline}).

\textbf{Embedding and Retrieval --} Each histopathology image is encoded into a feature vector using a vision foundation model. For a given query image, cosine similarity is computed against database embeddings, and the top-3 most similar images are retrieved. To prevent trivial matches, the query image itself is excluded from the retrieval index during evaluation.

We evaluate multiple retrieval backbones. UNI2 is a large-scale pathology foundation model designed to produce robust, semantically meaningful representations~\cite{chen2024uni2}. We further compare against CONCH, Virchow, KimiaNet, and PathDINO to assess how the structure of the embedding space influences downstream caption summarization. This comparison goes beyond architectural differences, directly testing whether improved pathology representations lead to more grounded and clinically reliable language generation.

\textbf{Caption Aggregation --}
The retrieved images are paired with their expert annotations. Due to the prevalence of repeated morphology terms and recurring textbook-style descriptions across related cases, the retrieval stage often provides clinically relevant language that would be difficult to generate reliably from image pixels alone. This is particularly important in pathology, where much of the semantic content lies not in generic object categories but in subtle descriptors of architecture, cytology, spatial arrangement, and contextual patterns.

\textbf{LLM Summarization --}
A large language model summarizes the retrieved captions into a concise final description. In this setting, the model functions as a \emph{summarizer} rather than a free generator. We employ Phi-4 and Qwen for this task~\cite{abdin2024phi4, bai2023qwen}. This distinction is central: instead of hallucinating content directly from the image, the model integrates expert-derived evidence into a compact, coherent output. In effect, the language model operates as a semantic compressor over retrieved pathology text.

\section{Experimental Setup}

\textbf{Dataset --} We evaluate on the ARCH histopathology dataset, which contains 7,563 images paired with expert-provided captions~\cite{gamper2021arch}. The dataset exhibits a vision–language, multiple-instance structure, where captions—drawn from textbooks and PubMed articles—encode dense clinical information, including morphology, diagnosis, and cross-image comparisons. This heterogeneous mix of didactic and scientific text, combined with variation in tissue types, staining, and magnification, makes ARCH well suited for retrieval-guided evaluation due to its rich domain-specific terminology. At the same time, ambiguous image–text alignment and complex medical language make it inherently challenging, highlighting the importance of semantic retrieval.

\textbf{Compared Models --} 
We compare retrieval-guided captioning against MedGemma, which generates captions directly from images~\cite{sellergren2025medgemma}. For retrieval-based experiments, we evaluate multiple visual backbones, including UNI/UNI2~\cite{chen2023uni}, CONCH~\cite{lu2024conch}, Virchow2~\cite{zimmermann2024virchow2}, and PathDINO~\cite{alfasly2023pathdino}, with KimiaNet included as a strong non-transformer baseline~\cite{riasatian2021kimianet}. This spectrum enables a controlled comparison between a pure image-to-text generator and retrieval-guided systems that differ only in the embedding space used for nearest-neighbor search.

\textbf{Evaluation Metric --}
Semantic alignment is quantified using cosine similarity within a BioBERT-based embedding space~\cite{lee2020biobert}. This approach is favored over exact lexical matching, as clinically valid pathology descriptions may exhibit substantial variation in wording while preserving equivalent meaning. In the domain of pathology, synonymy, diagnostic paraphrasing, and terminology compression are prevalent; accordingly, semantic similarity constitutes a more faithful and clinically relevant measure of caption quality than surface-level lexical overlap.

\textbf{Expert Evaluation Protocol --}
To assess clinical relevance, we conducted a single-blind expert evaluation involving one pathologist across 20 randomly selected cases. For each case, the reviewer was presented with two generated captions—one produced by MedGemma and one by the proposed retrieval-guided method—without disclosure of their respective sources. Diagnostic labels were intentionally withheld to ensure that evaluations were based solely on the visual content of the image. Assessment criteria included clinical plausibility, morphological fidelity, and descriptive specificity. This design was intended to mitigate confirmation bias and more closely approximate unbiased expert judgment.

\section{Results}

\textbf{Quantitative Comparison --}
Table~\ref{tab:main_results} summarizes the primary quantitative results on the ARCH dataset. MedGemma achieves a mean cosine similarity of 0.4732. In contrast, retrieval-guided captioning attains substantially higher performance, with the best configuration (UNI2) achieving a mean cosine similarity of 0.6039. 
The magnitude of the improvement is notable. The 95\% confidence intervals for UNI2 and MedGemma do not overlap, indicating that the observed gain is robust at the dataset level rather than driven by a limited subset of favorable cases. This finding suggests that the advantage conferred by retrieval grounding is systematic. 

\begin{table}[ht]
\centering
\caption{Semantic similarity (BioBERT cosine) between retrieval-guided generation (RGG) captions and ground-truth captions on the ARCH dataset. MedGemma serves as a baseline, generating captions solely from the input image without retrieval support.}
\label{tab:main_results}
\begin{tabular}{l c c c}
\toprule
Model & Cosine & 95\% CI & $\Delta$ vs MedGemma \\
\midrule
\textbf{MedGemma}  & 0.4732 & [0.4705, 0.4760] & -- \\
\midrule
RGG via UNI2            & \textbf{0.6039} & [0.6005, 0.6074] & \textbf{+0.1307} \\
RGG via CONCH$\dagger$                 & 0.5176 & [0.5139, 0.5212] & \textbf{+0.0444} \\
RGG via PathDINO  & 0.4999 & [0.4960, 0.5040] & \textbf{+0.0267} \\
RGG via KimiaNet              & 0.4500 & [0.4464, 0.4536] & -0.0232 \\
RGG via Virchow               & 0.4374 & [0.4338, 0.4409] & -0.0358 \\
\bottomrule
\multicolumn{4}{l}{ $\dagger$  CONCH can generate captions on its own but its decoder is not publicly available.}\\

\end{tabular}
\end{table}

\begin{figure}[ht]
\centering
\includegraphics[width=0.85\linewidth]{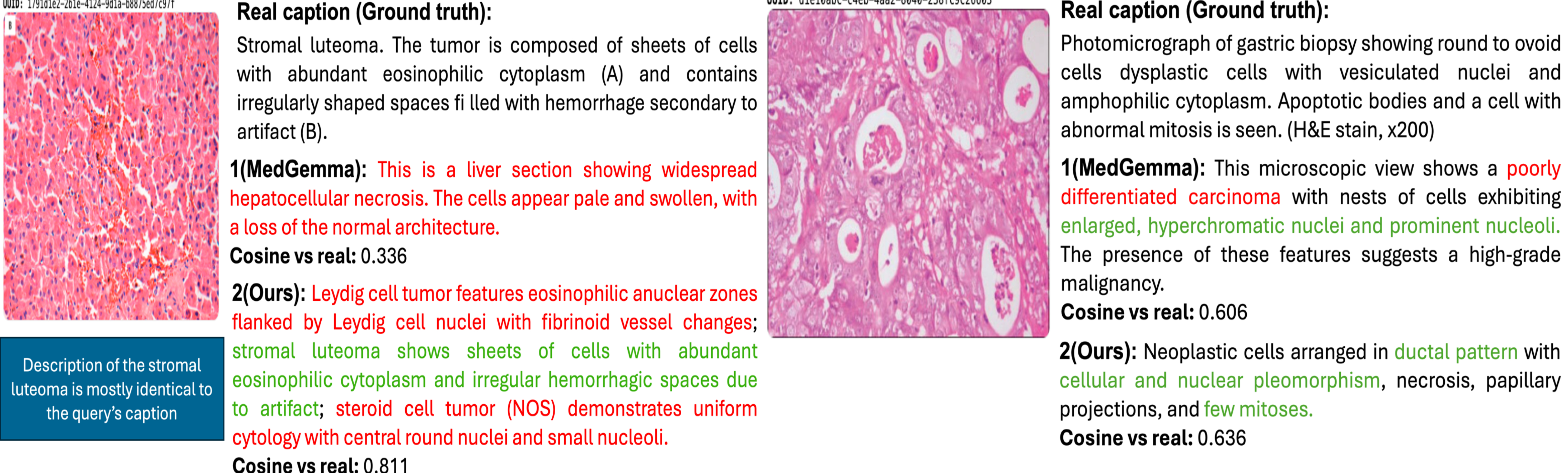}
\caption{Example of successful retrieval-guided generation (RGG). The generated summary exhibits stronger semantic alignment with the ground-truth  and more accurately captures morphology-relevant terminology (red: wrong; green: correct; blue box: pathologist comment).}
\label{fig:success_case}
\end{figure}

\begin{figure}[ht]
\centering
\includegraphics[width=0.85\linewidth]{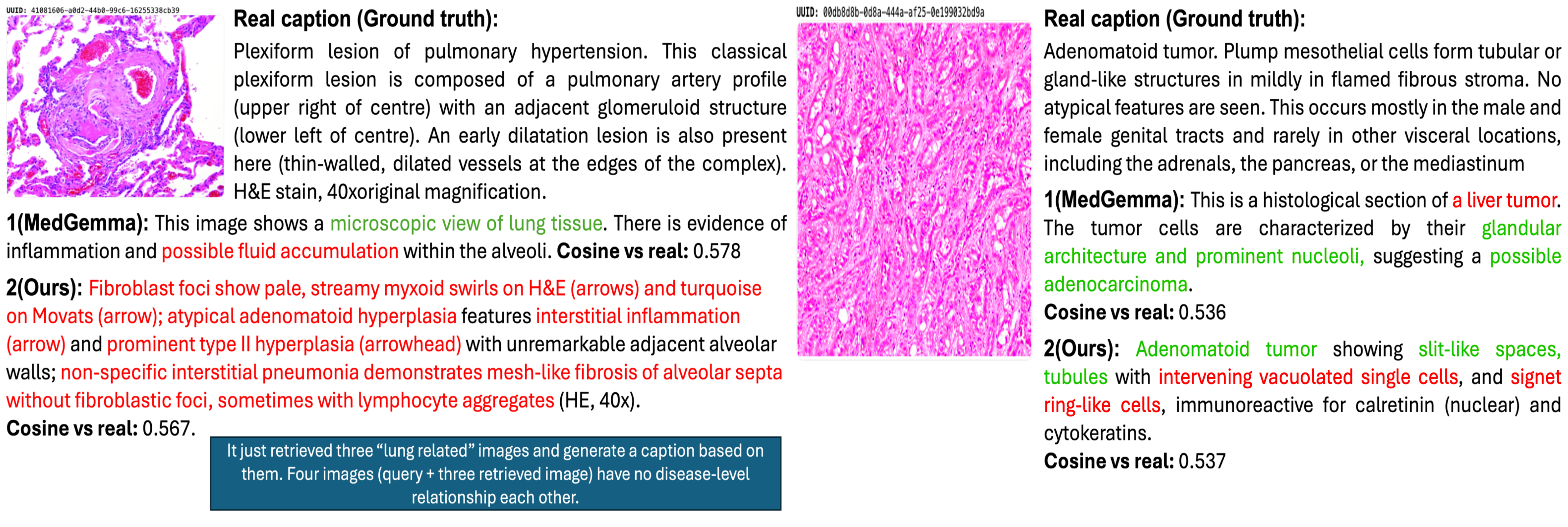}
\caption{Failure case: Concept mixing across unrelated or insufficiently matched diseases. Although the generated output is fluent, the description is clinically inconsistent reflecting a breakdown in retrieval alignment (red: wrong; green: correct; blue box: pathologist comment).}
\label{fig:failure_case}
\end{figure}

\textbf{Multi-Model Analysis --}
The comparative results reveal three distinct performance tiers. First, UNI2 clearly outperforms all other methods, indicating that stronger pathology-specific representations substantially enhance retrieval quality and, consequently, caption summarization. Second, CONCH and PathDINO exceed the MedGemma baseline, albeit with more modest gains. Third, KimiaNet and Virchow do not surpass the direct generative baseline on this task. This stratification underscores a central dependency: the effectiveness of retrieval-guided generation is critically determined by the quality of the visual embedding space. The approach is only as reliable as the relevance of the retrieved evidence it provides to the language model. When the retrieval backbone is well aligned with pathology semantics, as in UNI2, the resulting summaries exhibit markedly improved grounding and coherence.

\textbf{Results Interpretation --}
These findings support two primary conclusions. First, grounding generation in retrieved expert-authored captions improves semantic fidelity relative to direct image-only generation. Second, the choice of retrieval backbone constitutes a critical design factor. Pathology-specific representations not only enhance retrieval performance but also improve the quality of the evidence available for downstream generation. More broadly, retrieval-guided captioning may serve as a practical downstream benchmark for evaluating pathology foundation models. Improvements in representation quality translate into more coherent retrieval and more faithful summarization, extending beyond conventional retrieval-based evaluation metrics.

\section{Qualitative Analysis and Pathologist Review}
\begin{figure}[ht]
\centering
\includegraphics[width=0.85\linewidth]{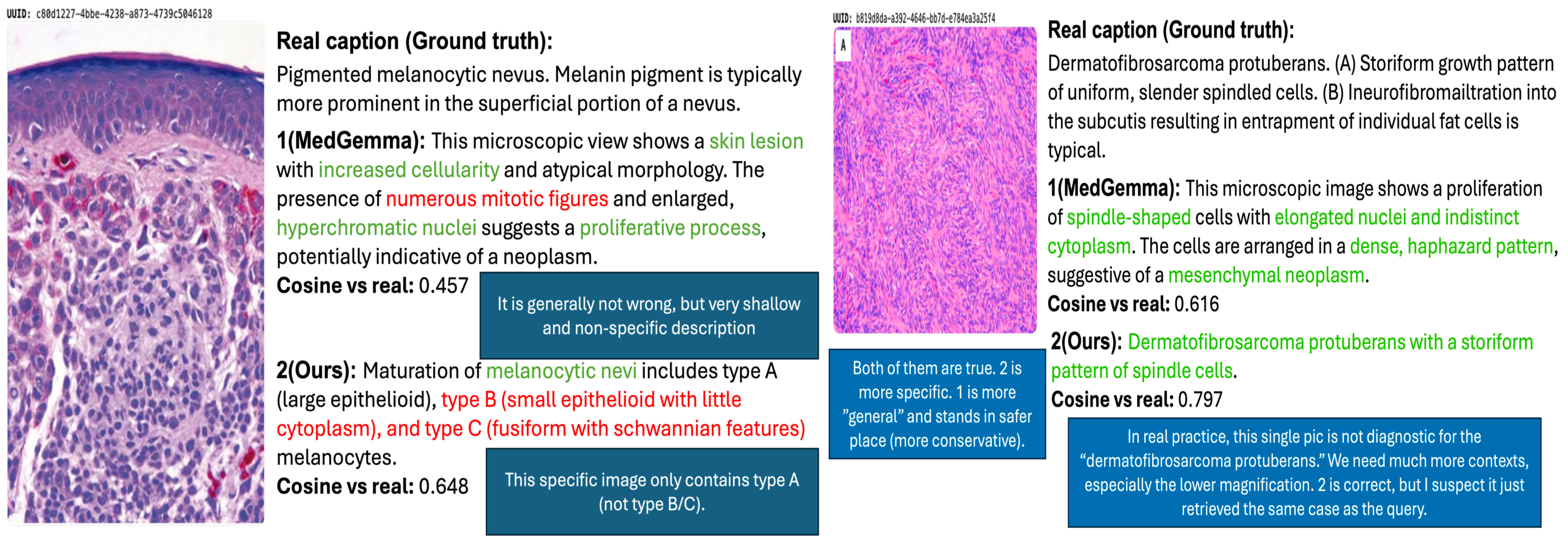}
\caption{Ambiguous case: the generated caption is partly accurate but shows unwarranted diagnostic specificity (red: wrong; green: correct; blue box: pathologist comment).}
\label{fig:ambiguous_case}
\end{figure}

Across the 20 reviewed cases, the pathologist consistently found retrieval-guided captions to be more morphology-aware, while also identifying important limitations, most notably concept mixing. In \textbf{successful} examples (Fig.~\ref{fig:success_case}), retrieval-guided captioning preserves key pathology terminology and maintains appropriate morphological structure. In multiple instances, these captions were preferred over MedGemma outputs due to their retention of descriptive morphological language, including terms related to cellular morphology, tissue architecture, and disease-associated structural features, which were often absent or attenuated in the direct generative baseline. A principal \textbf{failure} mode is \emph{concept mixing} (Fig.~\ref{fig:failure_case}), wherein the retrieved set contains visually related but diagnostically incongruent cases, and the summarization process combines these into a single, unified description. This limitation highlights a fundamental challenge in evidence aggregation: when the top retrieved examples are only weakly aligned at the semantic level, the summarizer may produce a clinically inconsistent caption. Such outputs can remain fluent and superficially plausible while being pathologically incorrect. Additionally, in some cases the retrieval-guided summary introduces a \textbf{highly specific} diagnostic label despite insufficient visual evidence to support such certainty (Fig.~\ref{fig:ambiguous_case}). This observation underscores the need for mechanisms to regulate diagnostic specificity in retrieval-guided generation.

The pathologist review indicates that these methods exhibit distinct risk profiles. MedGemma outputs tend to be more generic and, in some cases, anatomically inaccurate, whereas retrieval-guided generation (RGG) produces more specific and morphology-aware descriptions but may occasionally exhibit overcommitment. These findings are consistent with the quantitative model ranking, suggesting that stronger retrieval backbones may reduce the likelihood of such semantic drift.

\section{Discussions and Conclusion}

Retrieval improves performance by constraining the language model to summarize grounded expert text rather than generate unsupported content. The strong results of UNI2 likely stem from better semantic structure in its embedding space. More broadly, retrieval-guided captioning (RGG) reframes the task as evidence-grounded language modeling: the visual encoder retrieves relevant prior cases, while the language model compresses this evidence into a coherent caption. This separation appears to enhance trustworthiness compared to direct image-to-text generation, where both perception and clinical language must be inferred simultaneously. Additionally, repeated or near-repeated caption structures in ARCH should not be viewed purely as a limitation; they reflect the reality of pathology references, where multiple images map to shared disease concepts and standardized descriptions. However, important limitations remain. Semantic similarity does not fully capture clinical correctness, and evaluation was limited to a single pathologist reviewing 20 cases. RGG is also inherently dependent on retrieval quality—poor retrieval can still yield fluent but misleading outputs. Future work should include multi-reader studies, improved retrieval diversity, confidence-aware summarization, and evaluation aligned with clinical utility. In conclusion, we presented a retrieval-guided generation (RGG) framework for histopathology captioning and demonstrated that it improves semantic alignment over direct generation. On the ARCH dataset, UNI2 + RGG achieved a mean cosine similarity of 0.6039, suggesting a more grounded and reliable direction for pathology image description.

\bibliography{references}

\end{document}